\pdfoutput=1

\documentclass[11pt]{article}

\usepackage{ACL2023}

\usepackage{times}
\usepackage{latexsym}

\usepackage[T1]{fontenc}

\usepackage[utf8]{inputenc}

\usepackage{microtype}

\usepackage{inconsolata}
\usepackage{amssymb}
\usepackage{amsmath}
\usepackage{makecell}
\newcommand{\e}[1]{{\em #1}}

\title{Synthetic Dataset for Evaluating Complex Compositional Knowledge for Natural Language Inference}
\author{Sushma Anand Akoju, Robert Vacareanu, Haris Riaz, Eduardo Blanco, Mihai Surdeanu \\
University of Arizona  \\
\texttt{\{sushmaakoju, rvacareanu, hriaz, eduardoblanco, msurdeanu\}@arizona.edu} \\}
\begin{document}
\maketitle
\begin{abstract}
We introduce a synthetic dataset called Sentences Involving \e{Complex} Compositional Knowledge (SICCK) and a novel analysis that investigates the performance of Natural Language Inference (NLI) models to understand compositionality in logic. We produce 1,304 sentence pairs by modifying 15 examples from the SICK dataset \citep{marellietal2014sick}. To this end, we modify the original texts using a set of phrases -- modifiers that correspond to universal quantifiers, existential quantifiers, negation, and other concept modifiers in Natural Logic (NL) \citep{maccartney2009natural}. We use these phrases to modify the subject, verb, and object parts of the premise and hypothesis. Lastly, we annotate these modified texts with the corresponding entailment labels following NL rules. We conduct a preliminary verification of how well the change in the structural and semantic composition is captured by neural NLI models, in both zero-shot and fine-tuned scenarios. We found that the performance of NLI models under the zero-shot setting is poor, especially for modified sentences with negation and existential quantifiers. After fine-tuning this dataset, we observe that models continue to perform poorly over negation, existential and universal modifiers. 

\end{abstract}

\section{Introduction}
Natural language inference (NLI) has made tremendous progress in recent years, both in terms of datasets, e.g., SNLI \citep{bowman2015snli}, MultiNLI \cite{N181101}, Adversarial NLI \citep{nie2019adversarial}, NLI\_XY \citep{rozanova2021decomposing}, MonaLog \citep{hu2019monalog}, and methods \citep{yang2020xlnet,lan2020albert,wang2021entailment,wang2021infobert,devlinetal2019bert}.
However, many of these directions lack explainability, a critical drawback that limits their applicability to critical domains such as medical, legal, or financial. In contrast, Natural Logic (NL) \citep{maccartney2009natural} provides the necessary explainability through explicit {\em compositionality} that is driven by several relations that serve as building blocks (Forward Entailment (FE), Reverse Entailment (RE), Negation, Cover, Alternation, Equivalence, and Independence) as well as rules to combine them, which model changes in monotonicity. 

In this work, we analyze how well transformer networks trained for NLI understand the atomic reasoning blocks defined in NL, and how well they can compose them to detect changes in monotonicity \citep{Richardson_Hu_Moss_Sabharwal_2020,joshi2020taxinli}. 
To this end, we create a dataset containing 1304 sentences by modifying 15 premise/hypothesis pairs from the SICK dataset \cite{marellietal2014sick}.
The dataset is generated by modifying the premise and hypothesis sentences selected, as follows:
\begin{itemize}
    \item We append a series of modifiers to subject/verb/objects in the hypothesis/premise pairs. These modifiers include universal quantifiers (e.g., \e{every}, \e{always}), existential quantifiers (e.g., \e{some}, \e{at least}), negation, and adverbs/adjectives (e.g., \e{happy}, \e{sad}). Table~\ref{tab:mods} lists the complete set of modifiers used.
    \item We store the adjusted entailment label for each modifier pair to understand the shift in meaning from word-level changes within sentential contexts. More formally, we used the seven entailment relations as defined in \citep{maccartney2009natural}. These labels were generated manually for each example by following monotonicity calculus and natural logic.  For example, consider the premise: \e{an old man is sitting in a field} and the hypothesis: \e{a man is sitting in a field}, with the original SICK label: Forward Entailment. After adding the universal quantifier \e{every} to the aforementioned SICK example, the modified premise: \e{an old man is sitting in a field} and the original hypothesis: \e{\textbf{every} man is sitting in a field} are annotated with the adjusted label: Reverse Entailment. 
    
\end{itemize}



Using this dataset, we analyzed the capacity of three different NLI methods to correctly capture the change in entailment given the modified texts. In particular, the contributions of this work are as follows:
 \begin{enumerate}
    \item We propose a mechanism to generate synthetic data for NLI that enforces compositionality in reasoning. Following this mechanism, we produce 1,304 examples from 15 SICK \citep{marellietal2014sick} premise, hypothesis sentence pairs by modifying the sentences for subject, verb, and object respectively with a series of modifiers.
    The resulting dataset is freely available at \url{https://github.com/clulab/releases/tree/sushma/acl2023-nlrse-sicck}.
    \item We define specific annotation guidelines based on monotonicity calculus and natural logic \citep{maccartney2009natural} for annotating the modified premise and hypothesis sentences in the dataset above. The resulting labels are included in the dataset.
    \item We conducted an analysis to understand how well these structural and compositional changes are captured by neural NLI models, in both zero-shot and fine-tuned scenarios. Our analysis indicates that NLI models perform poorly over negation and several types of quantifiers. Fine-tuned NLI models do not show significant improvement in learning about compositional changes when compared to their zero-shot equivalent models over our dataset. This suggests that compositionality in reasoning remains a challenge for neural models of language.
    
 \end{enumerate}

\section{Related Work}

Natural Logic (NL) is a formal reasoning approach that makes use of syntactic structure and semantic properties of lexical items to understand compositionally \cite{maccartney2009natural}.

Logical reasoning is a known challenge for neural NLI models \citep{ravichander2019equate}. In particular, NLI models struggle to understand quantifiers, which is highlighted by the fact that these models do not generalize well over quantifier-driven inference tasks \citep{haruta2020logical}. The monotonicity calculus over quantifiers with token-level polarity has been explored using the CCG parser over the SICK dataset to generate a synthetic dataset that considers compositional data augmentation \citep{marellietal2014sick} and monotonicity calculus \citep{hu2019monalog}. Other recent research focused on language structures to highlight the importance of compositionality, i.e., the premise and hypothesis differ only in the order of the words, or the presence of antonyms, synonyms, or negation \citep{dasgupta2018evaluating}. Having such data augmentation can help move closer to the compositional encoding of the language \citep{dasgupta2018evaluating}. 
Our work extends this direction: our dataset captures both phrasal changes (e.g., synonyms, hypernyms), which we inherit from the SICK dataset \cite{marellietal2014sick}, as well as multiple types of modifiers that are critical for NLI such as universal, existential, negation, and adjectives/adverbs.

The FraCas test suite \cite{cooper1996using} contains 346 examples that explore aspects of natural logic applied to NLI~\cite{maccartney2009natural}. The HELP dataset \cite{yanaka2019help} modifies phrases in premise/hypothesis sentences based on monotonicity reasoning from combinatorial categorical grammar \cite{steedman2011combinatory} and semantic tagging \cite{abzianidze2017towards}. As mentioned above, our work is complementary to such datasets, as we cover other types of text modifications. The MED dataset \cite{yanakaetal2019neural} is another manually-labeled dataset where hypotheses were also modified by the human labelers given the monotonicity information for the premises. Similarly, we manually labeled NLI information, but our work focuses mainly on compositional information in a sentential context.

Enhancing the dataset with data augmentation is another recent method to test the generalizability of NLI models \citep{jha2020does}. Lexical entailment acquired from the distributional behavior of word pairs \citep{geffet2005distributional} led to the subsequent work of \cite{bowmanetal2015large}, who produced a 3-way classification task for NLI dataset that serves as a benchmark  for evaluating natural language understanding. Using Natural Logic as a means to learn and reason about the semantic and lexical relations is a common method used to improve the reasoning capabilities of the NLI models \citep{bowman2015recursive}. 

The NLI\_XY dataset \citep{rozanova2021decomposing} conducts structural investigation over the transformer-based NLI models. 
In particular, the authors investigate how monotonicity (upwards or downwards) changes when the premises and hypotheses are modified through the insertion of hypernym/hyponym phrases. This work is complementary to ours: while they focus on monotonicity in lexicalization (e.g., changing from a hypernym to a hyponym), we focus on changes in monotonicity due to explicit modifiers applied on top of such lexical modifications.

The MonaLog system \citep{hu2019monalog} introduces a simple yet explainable NLI method that relies on a simplified Natural Logic implementation. The proposed method operates by implementing monotonicity calculus over CCG syntactic trees using ``a small
inventory of monotonicity facts about quantifiers,
lexical items and token-level polarity.'' Despite its simplicity, the authors report excellent performance on the SICK dataset. More closely related to our work, they use MonaLog to generate additional training data for NLI from the generated proofs. 

\section{Dataset}
\label{sec:data}
We introduce a synthetic dataset to facilitate the analysis of compositionality in logic. 
The dataset contains 1,304 sentences that were created by modifying 15 examples from the SICK dataset \citep{marellietal2014sick} with a variety of modifiers.
To this end, we used a set of phrases that correspond to universal quantifiers, existential quantifiers, negation, and other concept modifiers in Natural Logic (NL) \citep{maccartney2009natural}. 
These modifiers were applied to syntactic constructs in both premise and hypothesis and the entailment labels are adjusted, as detailed below.

\subsection{Overview}

At a high level, our dataset creation followed the following steps:
\begin{enumerate}
    \item We start with 15 seed pairs of premise and hypothesis sentences from SICK. Table~\ref{tab:seeds} shows the seed sentence pairs.
    \item We syntactically analyze these sentences to understand their subject-verb-object (SVO) structures. Each of the SVO elements is then modified using a subset of the applicable modifiers listed in Table~\ref{tab:mods}. This process is detailed in Section~\ref{sec:sentmod}.
    \item Lastly, we re-annotate the entailment labels for the modified sentences, using the seven entailment relations defined in \citep{maccartney2009natural}: Forward Entailment (FE), Reverse Entailment (RE), Negation (Neg) (or Contradiction), Alternation, Cover, Independence (Neutral) and Equivalence (Equiv). This step is detailed in Section~\ref{sec:anno}. The labels are described in Table~\ref{tab:relations}.
\end{enumerate}

\begin{table*}[h]
\centering
\begin{small}
\begin{tabular}{lll}
\hline
\textbf{Premise} & \textbf{Hypothesis} & \textbf{SICK label} \\
\hline
an \textbf{old man} is sitting in a field & a \textbf{man} is sitting in a field & Entailment \\
A boy is standing in the \textbf{cold water}	& A boy is standing in the \textbf{water}	& Entailment\\
Two children are \textbf{hanging on a large branch} & Two children are \textbf{climbing a tree}	&	Entailment\\
A \textbf{boy} is hitting a baseball & A \textbf{child} is hitting a baseball	&	Entailment\\
Two dogs are playing by a \textbf{tree} &  Two dogs are playing by a \textbf{plant}	& Entailment\\
A player is \textbf{throwing the ball}	& \textbf{Two teams are competing in a football match}&	Neutral\\
A man is \textbf{sitting} in a field & A man is \textbf{running} in a field	&	Neutral\\
Two dogs are \textbf{playing} by a tree & Two dogs are \textbf{sleeping} by a tree	&	Neutral\\
\textbf{A girl with a black bag is on a crowded train} & \textbf{A cramped black train is on the bag of a girl}	&	Neutral\\
A blond girl is \textbf{riding the waves} & A blond girl is \textbf{looking at the waves}	&	Neutral\\
The \textbf{turtle} is following the \textbf{fish} & The \textbf{fish} is following the \textbf{turtle}	&	Contradiction\\
A man is jumping into an \textbf{empty} pool	& A man is jumping into a \textbf{full} pool	& Contradiction\\
A deer \textbf{is jumping} over a fence & A deer \textbf{isn't jumping} over the fence	&	Contradiction\\
A child is \textbf{hitting} a baseball	& A child is \textbf{missing} a baseball	&	Contradiction\\
A classroom is \textbf{full of students} & A classroom is \textbf{empty} &	Contradiction\\
\hline
\end{tabular}
\end{small}
\caption{15 premise/hypothesis sentence pairs from the SICK dataset \citep{marellietal2014sick} and corresponding NLI labels that form the seed of our dataset. The bold text highlights the lexically-driven compositional change in the premise and hypothesis sentences.}
\label{tab:seeds}

\end{table*}

\begin{table*}[h]
\centering
\begin{tabular}{lc}
\hline
\textbf{Modifier Type} & \textbf{Modifiers}\\
\hline
Universal quantifiers & \e{every, always, never, every one of}\\
Existential quantifiers & \e{some, at least, exactly one, all but one}\\
Negation & \e{not every, no, not} \\
Adjectives & \e{green, happy, sad, good, bad, an abnormal,} and \e{an elegant}\\
Adverbs & \e{abnormally, elegantly}\\
\hline
\end{tabular}
\caption{List of modifiers used to modify subject, verb, and object elements of sentences. They are applied to each of the premise and hypothesis sentences in Table 1.}
\label{tab:mods}
\end{table*}

\begin{table*}
\centering
\begin{tabular}{llp{60mm}}
\hline
\textbf{Entailment Relation} & \textbf{Set Theoretic Notation} & \textbf{Examples using WordNet Hierarchy}\\
\hline
Equivalence & $X \equiv Y $ & \e{couch} $\equiv $ \e{sofa} \\
Forward Entailment (FE) & $ X \subset Y$ & Hyponym: \e{crow} $\subset $  bird.\\
Reverse Entailment (RE) & $ X \supseteq Y$ & Hypernym: \e{Asian} $\supseteq$ \e{Thai}. \\
Negation (Neg) & $ {X \cap Y = \phi} \wedge {X \cup Y = \mathbb U}$ & Antonym: \e{able} $\neg $ \e{unable} \\
Alternation &  $ {X \cap Y = \phi} \wedge {X \cup Y \neq \mathbb U}$ & Typically caused concepts with a shared hypernym: \e{cat} $\|$ \e{dog}. The corresponding hierarchy is : \e{carnivore} $\rightarrow$ \e{feline}, \e{canine}; \e{feline} $\rightarrow$ \e{cat}; and \e{canine} $\rightarrow$ \e{dog}. \\
Cover & $ {X \cap Y \neq \phi} \wedge {X \cup Y = \mathbb U}$ & \e{animal} $ \smile$ \e{non-ape}\\
Independence & \makecell{ all other cases } & \e{hungry} $\|$ \e{hippo} \\
\hline
\end{tabular}
\caption{Entailment relations as defined in \citep{maccartney2009natural} with explanations using the WordNet hierarchy \citep{miller1995wordnet}.}
\label{tab:relations}
\end{table*} 


\subsection{Sentence Modification Strategy}
\label{sec:sentmod}

For each premise and hypothesis sentence pair, we modified
individual subject, verb, and object phrases with the following approach:
\begin{enumerate}
    \item To modify {\em subjects}, we used the Berkeley Neural Parser to extract the left-most noun phrases (NPs). We then append the applicable modifiers from Table~\ref{tab:mods}. In particular, we used universal quantifiers, existential quantifiers, negations, and adjectives. 
    
    \item To modify \emph{verbs}, we used the Berkeley Neural parser to extract the rightmost verb phrases (VPs) from the parse tree and appended the applicable modifiers. Verbs were modified using universal quantifiers (\e{always}, \e{never}), negations (\e{not}, \e{never}), and  adverbs (\e{abnormally},\e{elegantly}). 
    
    
    \item To detect \emph{objects}, we used the syntactic dependency parser of \cite{vacareanu-etal-2020-parsing} to identify noun phrases attached to the main verb. 
    Similarly to the subject modifications, these objects were modified using universal quantifiers, existential quantifiers, negations, and adjectives.
\end{enumerate}
After modifying each of the premises and hypotheses sentences, we generate multiple new data points as follows:
$f(P_i,H_i, m, SVO) = {{P_i}^{'},{H_i}^{'}}$ where $m \in M:$ all modifiers; $SVO:$ subject/verb/object phrases for either one of the parts of the sentence; and $P_i, H_i$ are premise and hypothesis from sentence pairs $S_i \in S$ where $S$ is the set of 15 examples from SICK. Lastly, $f$ is the function that modifies a given premise and hypothesis that follows one of the modification strategies described above. 
We generate the following pairs of combinations of the premise, and hypothesis sentences: 
${({P_i}^{'}, H_i), (P_i,{H_i}^{'}), ({P_i}^{'},{H_i}^{'})}$. 
We repeat this process to modify each of the relevant sentence phrases, as well as a couple of combinations: subject, verb, object,   subject $+$ object, and verb $+$ object.

\begin{table*}
\centering
\begin{small}
\begin{tabular}{lllll}
\hline
\textbf{Premise} & \textbf{ Hypothesis} & \textbf{SVO} & \textbf{Modifier Type} & \textbf{Label} \\
\hline 
an old man is sitting in a field  & a man is sitting in a field &  None & None & FE \\
\textbf{every} old man is sitting in a field & a man is sitting in a field &  Subject & Universal & FE\\
an old man is sitting in a field & \textbf{every} man is sitting in a field & Subject & Universal & RE\\
an old man is \textbf{elegantly} sitting in a field & a man is \textbf{elegantly} sitting in a field & Verb & Adverb & FE\\
an old man is sitting in \textbf{every} field & a man is sitting in a field & Object & Universal & FE \\
an old man is sitting in a field & a man is sitting in \textbf{every} field & Object & Universal & Neutral\\

\hline
\end{tabular}
\end{small}
\caption{Premise, hypothesis examples where one or both of the premise and hypothesis were modified. The text in bold indicates the change from the original text. The \e{SVO} column indicates the part of the sentence that was modified: subject, verb, or object (SVO). The Modifier type indicates which type of modifier was used to modify the parts of sentences. The label is the Entailment relation annotated by the annotators over modified data. }
\label{tab:examples}
\end{table*} 



\subsection{Entailment Annotation Strategy}
\label{sec:anno}

To annotate our dataset,\footnote{The annotation guidelines we followed are detailed on this website \url{https://github.com/clulab/releases/tree/sushma/acl2023-nlrse-sicck/annotations-guidelines/NLI_annotation_task_guidelines.pdf}} we created a set of annotation guidelines that follow Natural Logic and monotonicity calculus \citep{maccartney2009natural}. 



In general, to produce entailment relations we used a set theoretic approach to understand how the set of concepts that holds true in the premise overlaps with the set described in the hypothesis. 
To implement this set theoretic approach consistently, we defined the quantitative interpretation for several more ambiguous modifiers such as \e{all but one}, \e{all}, \e{not every} as follows: 
\begin{enumerate}
    \item For the modifier \e{all}, we consider the size of the set of elements X to be greater than 0: $|X| > 0 $. For example, in the case of the phrase \e{all children}, we consider the size of the set of children to be greater than 0.
    \item For the \e{all but one} modifier, we consider the size of \e{all} as $N$ and the size of \e{all but one} to be $N-1$. 
    Note that the size of \e{all but one} could thus theoretically be 0, when $N = 1$.
    \item  For \e{not every} we consider the size of the corresponding set X to be 0 or larger: $|X| \geq 0$ where X is any set defined over the sentence. \e{not every man} would make X as a set of all men but there exists zero or one or more men that would not be included in this set.
    \item When we cannot determine the size of the intersection of the two sets of premise and hypothesis, we resolved the annotation to be a Neutral label among all 7 entailment relations.
    \item When comparing quantifiers between modified premise, and hypothesis sentence pairs, we denote the sizes of sets mathematically for $P \cup H$, $P \cap H$, and the Universal set.  For example, consider the premise: \e{\textbf{every} turtle is following the fish} and the hypothesis: \e{\textbf{every} fish is following the turtle}. The set over the premise is $P: \forall X \in \textrm{all turtles following one fish}$, and the set over hypothesis is $H: \forall X \in \textrm{all fishes following the turtle} $. 
    Thus, $P \cap H = \phi$. In this case, the label is Negation (see Table 3).  Table~\ref{tab:examples} includes more examples with the corresponding entailment labels.
    \end{enumerate}

A total of 1,304 modified premise and hypothesis sentence pairs along with original sentence pairs were included in the final SICCK dataset. The data was annotated by 5 annotators which were distributed between two sub-groups of annotators, based on the complexity of the labels. 
In the first two rounds of annotations, we re-grouped to develop concrete guidelines for annotations, without defining too strict rules by leaving room for more natural ``if-this-then-that'' deductions.
There were disagreements between annotations which were resolved by verifying the sizes of sets mathematically over $X \cup Y$, $X \cap Y$ to follow the entailment relations defined as in \citep{maccartney2009natural}. While in the initial round the inter-annotator agreement was low ($k < 0.4$), the annotations were revised until each group of annotators converged. 

Tables~\ref{tab:statsmods},~\ref{tab:statssyn}, and~\ref{tab:statslab} provide summary statistics about the SICCK dataset.

\begin{table}[t]
\begin{tabular}{lc}
\hline
\textbf{Modifier type} & \textbf{\# of sentence pairs}\\
\hline
With universal quantifiers & 217\\
With existential quantifiers & 303 \\
With negation & 167\\
With adjectives/adverbs & 602\\
\hline
\end{tabular}
\caption{Sentence counts in SICCK based on types of modifiers.}
\label{tab:statsmods}
\end{table}

\begin{table}[t]
\begin{tabular}{lc}
\hline
\textbf{SVO modified} & \textbf{\# of sentence pairs}\\
\hline
Subject & 560\\
Verb & 220 \\
Object & 509\\
\hline
\end{tabular}
\caption{Sentence counts in SICCK based on which syntactic structures are modified.}
\label{tab:statssyn}
\end{table}


\begin{table}[t]
\begin{tabular}{lc}
\hline
\textbf{Entailment relations} & \textbf{\# of sentence pairs}\\
\hline
Forward Entailment & 223\\
Reverse Entailment & 27\\
Alternation & 121\\
Negation & 54\\
Negation|Alternation & 260 \\
Neutral & 393\\
Equivalence & 7\\
Cover & 1\\
Cover|FE & 1\\
\hline
\end{tabular}
\caption{Label counts in SICCK. Note that Negation|Alternation indicates ambiguous labels where the two annotators did not converge.}
\label{tab:statslab}
\end{table}

\section{Evaluation}
We conducted an evaluation of how NLI methods capture the explicit compositionality in our dataset using two configurations: a zero-shot setting, in which we used NLI systems trained externally, and a fine-tuned setting, in which the same models were fine-tuned using our dataset.

\subsection{Zero-shot Analysis of NLI Models}

For this analysis, we evaluate three pretrained neural entailment models on our dataset. However, all these systems emit just the three ``traditional'' entailment labels (Forward Entailment, Contradiction, and Neutral) whereas our dataset contains the seven labels from NL. To align these label spaces, we performed the following transformations:

\begin{enumerate}
    \item In case a system produces a Neutral label, we run the prediction in the opposite direction, i.e., from hypothesis to premise. If the top label in the reversed direction is Forward Entailment (FE), we label the pair as Reverse Entailment. Otherwise, we keep the Neutral label. This heuristic allows these systems to produce four labels instead of three.
    \item We convert our seven labels to four labels through the following  heuristics: (a) Equivalence was removed since we had only one sentence pair labeled as Equivalence in our dataset; (b) Alternation is merged with Negation; (c) Cover and Independence become Neutral; and (d) the 7 examples that were annotated as Cover|FE were removed.
\end{enumerate}


We conducted zero shot evaluation using three NLI models: the cross-encoder model of \citet{reimers2019sentencebert} ({\tt nli-deberta-v3-base} in our tables), the adversarial NLI model of \citet{nieetal2020adversarial} ({\tt ynie/roberta-large-\dots}), and ELMo-based Decomposable Attention model \citep{Parikh2016ADA} ({\tt pair-classification-\dots}). We draw the following observations from this experiment:


\begin{itemize}
    \item Table~\ref{tab:zero-overall} indicates that the ELMO-based NLI model performs considerably worse than the other two transformer-based models. This is a testament to how far our field has progressed in just a few years. However, no model approaches 70 F1 points, which indicates that none of these models truly understand the task well.

    \item The NLI models do better over adjectives and adverbs, but they struggle to understand statements modified with universal and existential quantifiers, and negation. Tables~\ref{tab:zero-overall}--\ref{tab:adj} indicates that the transformer-based NLI models perform at over 70 F1 points on adjectives/adverbs, at over 65 F1 for universal quantifiers, at approximately 60 F1 for existential quantifiers, and at only 30--35 F1 for negation. This is a surprising finding considering how much attention negation has received in the NLP literature \cite{pangetal2002thumbs} \cite{hossainetal2022analysis} \cite{hossainetal2020analysis}.

    \item Lastly, Tables~\ref{tab:zero-subj}--\ref{tab:zero-obj} indicate that NLI models process objects best, followed by subjects, and, lastly, verbs. This is not surprising considering the increased semantic ambiguity of verbs. 
\end{itemize}

\begin{table*}[h]
\begin{tabular}{lllll}
\hline
\textbf{\small NLI system} & \textbf{\small F1} & \textbf{\small Precision} & \textbf{\small Recall} & \textbf{\small Accuracy}\\
\hline
\small nli-deberta-v3-base & \small \textbf{0.5254} & \small 
 \textbf{0.5601} & \small \textbf{0.5860} & \small \textbf{0.6579}\\
\small ynie/roberta-large-snli-mnli-fever-anli-R1-R2-R3-nli & \small 0.5200 & \small 0.5156 & \small 0.5533 & \small 0.6334\\
\small pair-classification-decomposable-attention-elmo & \small 0.0829 & \small 0.0497 & \small 0.2500 & \small 0.1986\\
\hline
\end{tabular}
\caption{\small Overall scores for the three pretrained NLI modes under zero-shot setting, based on compressed 4-entailment \\
relations: Forward Entailment, Reverse Entailment, Contradiction, and Neutral. 
 } 
\label{tab:zero-overall}
\end{table*}
\subsection{Analysis of Fine-tuned NLI models}

To understand if NLI methods are capable of learning this compositional information, we fine-tuned the two NLI models that performed better over the SICCK dataset. To maximize the data available, we implemented a 5-fold cross-validation evaluation over the entire SICCK dataset and experimented with multiple hyperparameters. In particular, we used 4 or 8 epochs, and batch sizes of 8, 16, or 32 data points.
\begin{table*}[h]
\begin{tabular}{lllll}
\hline
\textbf{\small NLI model with epochs, batch size} & \textbf{\small F1} & \textbf{\small Precision} & \textbf{\small Recall} & \textbf{\small Accuracy}\\
\hline
        \small ynie/roberta-large-snli-mnli-fever-anli-R1-R2-R3-nli-4-8  & \small (0.52$\pm$0.02) & \small (0.54$\pm$0.03) & \small (0.52$\pm$0.03) & \small (0.65$\pm$0.04) \\  
        \small nli-deberta-v3-base-4-8 & \small (0.33$\pm$0.02) & \small (0.36$\pm$0.03) & \small (0.38$\pm$0.02) & \small (0.38$\pm$0.02) \\  
        \small ynie/roberta-large-snli-mnli-fever-anli-R1-R2-R3-nli-4-16 & \small (0.59$\pm$0.04) & \small (0.59$\pm$0.04) & \small (0.60$\pm$0.05) & \small (0.74$\pm$0.06) \\  
        \small nli-deberta-v3-base-4-16 & \small (0.34$\pm$0.01) & \small (0.38$\pm$0.01) & \small (0.39$\pm$0.02) & \small (0.39$\pm$0.02) \\  
        \small ynie/roberta-large-snli-mnli-fever-anli-R1-R2-R3-nli-4-32 & \small \textbf{(0.62$\pm$0.04)} & \small \textbf{(0.61$\pm$0.04)} & \small \textbf{(0.63$\pm$0.04)} & \small \textbf{(0.79$\pm$0.05)} \\  
        \small nli-deberta-v3-base-4-32 & \small (0.37$\pm$0.01) & \small (0.41$\pm$0.01) & \small (0.42$\pm$0.01) & \small (0.42$\pm$0.01) \\  
        \small ynie/roberta-large-snli-mnli-fever-anli-R1-R2-R3-nli-8-8 & \small (0.49$\pm$0.06) & \small (0.50$\pm$0.05) & \small (0.49$\pm$0.06) & \small (0.60$\pm$0.07) \\  
        \small nli-deberta-v3-base-8-8 & \small (0.33$\pm$0.02) & \small (0.37$\pm$0.02) & \small (0.38$\pm$0.01) & \small (0.38$\pm$0.01) \\  
        \small ynie/roberta-large-snli-mnli-fever-anli-R1-R2-R3-nli-8-16 & \small (0.53$\pm$0.04) & \small (0.54$\pm$0.03) & \small (0.55$\pm$0.04) & \small (0.66$\pm$0.06) \\  
        \small nli-deberta-v3-base-8-16 & \small (0.33$\pm$0.02) & \small (0.36$\pm$0.02) & \small (0.37$\pm$0.03) & \small (0.37$\pm$0.03) \\  
        \small ynie/roberta-large-snli-mnli-fever-anli-R1-R2-R3-nli-8-32 & \small (0.57$\pm$0.01) & \small (0.56$\pm$0.01) & \small (0.58$\pm$0.02) & \small (0.72$\pm$0.02) \\  
        \small nli-deberta-v3-base-8-32 & \small (0.34$\pm$0.01) & \small (0.38$\pm$0.02) & \small (0.38$\pm$0.02) & \small (0.38$\pm$0.02) \\    
\hline
\end{tabular}
\caption{\small Overall scores for two {\em fine-tuned} NLI models on SICCK dataset based on the compressed 4-entailment relations: \\
Forward Entailment, Reverse Entailment, Contradiction, and Neutral. We repeated these experiments 5 times with different \\
random seeds; we report averages and standard deviation, for 4 and 8 epochs and batch sizes of 8, 16, and 32 data points.}
\label{tab:finetuned}
\end{table*}

The results of these experiments are summarized in Table~\ref{tab:finetuned}.
We draw the following observation from this experiment:
\begin{itemize}
    \item The difference in F1 scores between the fine-tuned systems and the corresponding zero-shot setting ranges from -0.19 to 0.2.
    This indicates that these systems do not acquire substantial new knowledge despite the fact that they've been exposed to approximately 1,300 sentences with compositional information. This suggests that understanding compositionality is harder than expected.
    \item Similar to the zero-shot setting, NLI models did better over adjectives, and adverbs and relatively better over existential quantifiers in comparison to that of the negation and universal quantifiers. We also observed that models seem to be confused when the annotated label was Neutral but the modifier types were negations.
    \item NLI models perform somewhat better over subject and object-modified examples than on examples with modified verbs. This indicates that the semantic ambiguity of verbs is likely to impact NLI models. 
\end{itemize}

\section{Error Analysis}
We analyze the incorrect predictions of the NLI models over SICCK dataset in this section. 
We observed that NLI models performed better over adjectives and adverbs, and relatively well over universal quantifiers in comparison to sentences modified with negation and existential quantifiers under both fine-tuned as well as zero-shot settings. We also observed that models seem to be confused when the adjusted label was Neutral and the modifier types were negations.

\subsection{Neutral Labels with Negation Modifiers}
Negation understanding in natural language has been a challenging problem \cite{pangetal2002thumbs,hossainetal2022analysis}. \cite{hossainetal2022analysis} discussed that Negation is underrepresented in natural language corpora. Further, \cite{hossainetal2020analysis} show that even though the transformers were fine-tuned with modified premises with negation (i.e., verb modifiers with negation), the transformers struggle with inference over negated sentence pairs.

In our SICCK dataset, there are 167 examples with negation modifiers. Table~\ref{tab:negsvo} shows some statistics relevant to this.
\begin{table}[t]
\begin{center}
\begin{tabular}{lllc}
\hline
 SVO & \# count & Neutral\\
\hline
\e{subject} & 86 & 65\\
\e{verb} & 41 & 31\\
\e{object} & 40 & 22\\
\hline
\end{tabular}
\end{center}
\caption{For all our SICCK dataset's 167 examples with negation modifiers, this table includes counts of all the modified subject, verb, and object parts of sentences respectively for each of the 4-Entailment adjusted labels from SICCK annotations. The last column indicates how many of these data points have the Neutral label.} 
\label{tab:negsvo}
\end{table}
Of these 167 examples with Negation modifiers, there are 118 
Neutral examples. We observed that 
nli-deberta-v3-base model incorrectly predicted ground truth for approximately 70\% of these examples and while the other NLI model (ynie/roberta-large-snli-mnli-fever-anli-R1-R2-R3-nli) incorrectly predicted 23\% of the examples. For all the incorrectly predicted labels for negation-modified examples with Neutral labels, the models seemed to be confused for various compositional cases, i.e. subject or verb or object-modified examples almost equivalently. Modifiers such as \e{no}, \e{not every}, \e{not} with Neutral and Contradiction labels seem to contribute to the confusion. SICCK examples also include the format of alternating modifiers between premises, hypothesis, or both i.e. ${({P_i}^{'}, H_i), (P_i,{H_i}^{'}), ({P_i}^{'},{H_i}^{'})}$ Section~\ref{sec:data} which further seems to confuse the NLI models. This is surprising since we have 593 Neutral examples in our SICCK dataset, albeit with fewer negation examples. Since the dataset is small and has a limited number of examples with negation modifiers, the evaluation analysis seems less generalizable. As emphasized by the analysis from \cite{hossainetal2022analysis} and \cite{hossainetal2020analysis}, detecting negation in natural language continues to be an unresolved problem. 


%
%

\subsection{Verb-modified Examples}
For verb modifiers, we selected \e{abnormally}, \e{elegantly}, \e{always}, \e{never}. Our SICCK dataset has a total of 220 verb-modified examples of which, we have 89 universal modifiers, 90 adverbs/adjectives, and 41 negation. Among the 31 verb-modified examples with negation modifiers and with Neutral label, NLI models incorrectly alternate between Contradiction and FE for 99\% of the examples. Of the 49 examples with universal modifiers over verbs with Neutral labels, approximately 69.4\% were incorrectly predicted. This further emphasizes that negation (especially when occurring in Neutral examples) remains a challenge.

\section{Conclusion}

This paper introduced a new, synthetic dataset that facilitates analyses of how NLI models capture compositionality. The dataset contains 1,304 sentence pairs that were created by modifying 15 examples from the SICK dataset \citep{marellietal2014sick} with a variety of modifiers that correspond to universal quantifiers, existential quantifiers, negation, and other concept modifiers in Natural Logic (NL) \citep{maccartney2009natural}.  We used these phrases to modify the subject, verb, and object parts of the premise and hypothesis. Lastly, we annotated these modified texts with the corresponding entailment labels following NL rules. 

We conducted a preliminary analysis of how well the change in the structural and semantic composition is captured and detected by neural NLI models, in both zero-shot and fine-tuned scenarios. We found that the performance of NLI models is poor in both settings, especially for modified sentences with negation and existential quantifiers, and when verbs are modified. 

\section*{Limitations}
While this work explores the impact of the typical compositional modifiers on entailment relations, we did not consider other fine-grained information that further captures upward or downward monotonicity from the monotonicity calculus of the premise/hypothesis sentence pairs. Further, the dataset that we generated is relatively small, at approximately 1,300 sentences. We also did not evaluate the dataset over T5, BART, GPT-x, and other state-of-the-art LLMs, which may provide more insights.
We also did not conduct any evaluation for explanations and interpretation of the evaluated NLI models, which could be future work.
Lastly, we did not include a comparison with existing datasets that were created specifically for negation modifiers and universal \& existential quantifiers. We see all these issues as exciting avenues for future work.




\bibliography{anthology,custom}
\bibliographystyle{acl_natbib}

\appendix

\section{Appendix}
\label{sec:appendix}
We provide the modifier-based evaluation results for zero-shot setting over NLI models and finetuned NLI models.

\begin{table*}
\begin{tabular}{lllll}
\hline
\textbf{\small NLI system} & \textbf{\small F1} & \textbf{\small Precision} & \textbf{\small Recall} & \textbf{\small Accuracy}\\
\hline

\small nli-deberta-v3-base & \small 0.5333 & \small 0.5474 & \small 
 \textbf{0.5877} & \small 0.6636\\
\small ynie/roberta-large-snli-mnli-fever-anli-R1-R2-R3-nli & \small \textbf{0.5597} & \small \textbf{0.5636} & \small 0.5854 & \small 
 \textbf{0.6774}\\
\small pair-classification-decomposable-attention-elmo & \small 0.0694 & \small 0.0403 & \small 0.2500 & \small 0.1613\\
\hline
\end{tabular}
\caption{{\bf Universal quantifiers}: \small scores based on compressed 4-entailment relations for zero-shot NLI evaluation.}
\label{tab:zero-universal}
\end{table*}

\begin{table*}
\begin{tabular}{lllll}
\hline
\textbf{\small NLI system} & \textbf{\small F1} & \textbf{\small Precision} & \textbf{\small Recall} & \textbf{\small Accuracy}\\
\hline

\small nli-deberta-v3-base & \small  \textbf{0.5101} & \small \textbf{0.5453} & \small  \textbf{0.5608} & \small \textbf{0.6106}\\
\small ynie/roberta-large-snli-mnli-fever-anli-R1-R2-R3-nli & \small 0.4932 & \small 0.4888 & \small 0.5348 & \small 0.5941\\
\small pair-classification-decomposable-attention-elmo & \small 0.1044 & \small 0.0660 & \small 0.2500 & \small 0.2640 \\
\hline
\end{tabular}
\caption{{\bf Existential quantifiers}: \small scores based on compressed 4-entailment relations for zero-shot NLI evaluation.}
\label{tab:zero-exist}
\end{table*}

\begin{table*}
\begin{tabular}{lllll}
\hline
\textbf{\small NLI system} & \textbf{\small F1} & \textbf{\small Precision} & \textbf{\small Recall} & \textbf{\small Accuracy}\\
\hline

\small nli-deberta-v3-base & \small 0.2678 & \small \textbf{0.3556} & \small 0.4637 & \small \textbf{0.3054}\\
\small ynie/roberta-large-snli-mnli-fever-anli-R1-R2-R3-nli & \small \textbf{0.2996} & \small 0.3470 & \small \textbf{0.4690} & \small 0.3533\\
\small pair-classification-decomposable-attention-elmo & \small 0.0412 & \small 0.0225 & \small 0.2500 & \small 0.0898\\
\hline
\end{tabular}
\caption{{\bf Negation Modifiers }: \small scores based on compressed 4-entailment relations for zero-shot NLI evaluation. }
\label{tab:zero-neg}
\end{table*}

\begin{table*}
\begin{tabular}{lllll}
\hline
\textbf{\small NLI system} & \textbf{\small F1} & \textbf{\small Precision} & \textbf{\small Recall} & \textbf{\small Accuracy}\\
\hline

\small nli-deberta-v3-base & \small \textbf{0.5768} & \small \textbf{0.6088} & \small  \textbf{0.6154} & \small \textbf{0.7741}\\
ynie/roberta-large-snli-mnli-fever-anli-R1-R2-R3-nli & \small 0.5543 & \small 0.5618 & \small 0.5523 & \small 0.7126\\
pair-classification-decomposable-attention-elmo & \small 0.1139 & \small 0.0687 & \small 0.3333 & \small 0.2060\\
\hline
\end{tabular}
\caption{{\bf Adjectives/Adverbs Modifiers}:\small scores based on compressed 4-entailment relations for zero-shot NLI evaluation.}
\label{tab:adj}
\end{table*}

\begin{table*}
\begin{tabular}{lllll}
\hline
\textbf{\small NLI system} & \textbf{\small F1} & \textbf{\small Precision} & \textbf{\small Recall} & \textbf{\small Accuracy}\\
\hline

\small nli-deberta-v3-base & \small \textbf{0.5070} & \small \textbf{0.5469} & \small  \textbf{0.5732} & \small \textbf{0.6429} \\ \hline
\small ynie/roberta-large-snli\_mnli\_fever\_anli\_R1\_R2\_R3-nli & \small 0.4862 & \small 0.4812 & \small 0.5168 & \small 0.6054\\ \hline
\small pair-classification-decomposable-attention-elmo & \small 0.0796 & \small 0.0473 & \small 0.2500 & \small 0.1893 \\
\hline
\end{tabular}
\caption{{\bf Modified subject}: \small scores based on compressed 4-entailment relations for zero-shot NLI evaluation. }
\label{tab:zero-subj}
\end{table*}

\begin{table*}
\begin{tabular}{lllll}
\hline
\textbf{\small NLI system} & \textbf{\small F1} & \textbf{\small Precision} & \textbf{\small Recall} & \textbf{\small Accuracy}\\
\hline

\small nli-deberta-v3-base & \small \textbf{0.4724} & \small  \textbf{0.5141} & \small  \textbf{ 0.5875} & \small  \textbf{0.5727} \\ \hline
\small ynie/roberta-large-snli\_mnli\_fever\_anli\_R1\_R2\_R3-nli & \small  0.4932 & \small  0.5082 & \small  0.5764 & \small  0.5955 \\ \hline
\small pair-classification-decomposable-attention-elmo & \small  0.0669 & \small  0.0386 & \small  0.2500 & \small  0.1545\\
\hline
\end{tabular}
\caption{{\bf Modified verb}: \small scores based on compressed 4-entailment relations for zero-shot NLI evaluation.}
\label{tab:zero-verb}
\end{table*}

\begin{table*}
\begin{tabular}{lllll}
\hline
\textbf{\small NLI system} & \textbf{\small F1} & \textbf{\small Precision} & \textbf{\small Recall} & \textbf{\small Accuracy}\\
\hline

\small nli-deberta-v3-base & \small  \textbf{0.5683} & \small  \textbf{0.6166} & \small  \textbf{0.6030} & \small  \textbf{0.7073} \\ \hline
\small ynie/roberta-large-snli\_mnli\_fever\_anli\_R1\_R2\_R3-nli & \small  0.5687 & \small  0.5638 & \small  0.5908 & \small  0.6778 \\ \hline
\small pair-classification-decomposable-attention-elmo & \small  0.0915 & \small  0.0560 & \small  0.2500 & \small  0.2240 \\
\hline
\end{tabular}
\caption{{\bf Modified object}: \small scores based on compressed 4-entailment relations for zero-shot NLI evaluation.}
\label{tab:zero-obj}
\end{table*}

\begin{table*}
\begin{tabular}{lllll}
\hline
\textbf{\small NLI system} & \textbf{\small F1} & \textbf{\small Precision} & \textbf{\small Recall} & \textbf{\small Accuracy}\\
\hline

\small ynie/roberta-large-snli-mnli-fever-anli-R1-R2-R3-nli & \textbf{\small (0.53$\pm$0.02)} & \textbf{\small (0.52$\pm$0.02)} & \textbf{\small (0.57$\pm$0.03)} & \textbf{\small (0.66$\pm$0.04)} \\ 
\small nli-deberta-v3-base & \small (0.35$\pm$0.02) & \small (0.38$\pm$0.02) & \small (0.38$\pm$0.01) & \small (0.47 $\pm$ 0.01) \\ 
\hline
\end{tabular}
\caption{{\bf Universal quantifiers}: \small Fine-tuned NLI models' evaluation scores based on 4-entailment relations. \\
We repeated these experiments 5 times with different random seeds; we report averages and standard deviation, for  \\
8 epochs and a batch size of 32 data points.} 
\label{tab:finetuned-universal}
\end{table*}
\begin{table*}
\begin{tabular}{lllll}
\hline
\textbf{\small NLI system} & \textbf{\small F1} & \textbf{\small Precision} & \textbf{\small Recall} & \textbf{\small Accuracy}\\
\hline

        \small ynie/roberta-large-snli-mnli-fever-anli-R1-R2-R3-nli & \textbf{\small (0.59$\pm$0.03)} & \textbf{ \small (0.60$\pm$0.03)} & \textbf{\small (0.60$\pm$ 0.04)} & \textbf{\small (0.77$\pm$0.05)} \\ 
        \small nli-deberta-v3-base & \small (0.38$\pm$0.01) & \small (0.41$\pm$0.02) & \small (0.40$\pm$0.01) & \small (0.48$\pm$0.02) \\
\hline
\end{tabular}
\caption{{\bf Existential quantifiers}: \small Fine-tuned NLI models' evaluation scores based on 4-entailment relations. \\
We repeated these experiments 5 times with different random seeds; we report averages and standard deviation, for  \\
8 epochs and a batch size of 32 data points.}
\label{tab:finetuned-exist}
\end{table*}
\begin{table*}
\begin{tabular}{lllll}
\hline
\textbf{\small NLI system} & \textbf{\small F1} & \textbf{\small Precision} & \textbf{\small Recall} & \textbf{\small Accuracy}\\
\hline

\small ynie/roberta-large-snli-mnli-fever-anli-R1-R2-R3-nli & \textbf{\small (0.40$\pm$0.05)} & \textbf{\small (0.37$\pm$0.04)} & \textbf{\small (0.48 $\pm$0.06)} & \textbf{\small (0.61$\pm$0.04)} \\ 
\small nli-deberta-v3-base & \small(0.19$\pm$0.05) & \small(0.23$\pm$0.05) & \small(0.34$\pm$0.05) & \small(0.17$\pm$0.02) \\ 
\hline
\end{tabular}
\caption{{\bf Negation}: \small Fine-tuned NLI models' evaluation scores based on 4-entailment relations. \\
We repeated these experiments 5 times with different random seeds; we report averages and standard deviation, for  \\
8 epochs and a batch size of 32 data points.}
\label{tab:finetuned-neg}
\end{table*}
\begin{table*}
\begin{tabular}{lllll}
\hline
\textbf{\small NLI system} & \textbf{\small F1} & \textbf{\small Precision} & \textbf{\small Recall} & \textbf{\small Accuracy}\\
\hline

\small ynie/roberta-large-snli-mnli-fever-anli-R1-R2-R3-nli & \textbf{\small (0.58$\pm$0.02)} & \textbf{\small (0.59$\pm$0.02)} & \textbf{\small (0.58$\pm$0.02)} & \textbf{\small (0.75$\pm$0.03)} \\ 
\small nli-deberta-v3-base & \small (0.35$\pm$0.02) & \small (0.39$\pm$0.03) & \small (0.38$\pm$0.02) & \small (0.49$\pm$0.02) \\
\hline
\end{tabular}
\caption{{\bf Adjectives/Adverbs}: \small Fine-tuned NLI models' evaluation scores based on 4-entailment relations. \\
We repeated these experiments 5 times with different random seeds; we report averages and standard deviation, for  \\
8 epochs and a batch size of 32 data points.}
\label{tab:finetuned-adj}
\end{table*}
\begin{table*}
\begin{tabular}{lllll}
\hline
\textbf{\small NLI system} & \textbf{\small F1} & \textbf{\small Precision} & \textbf{\small Recall} & \textbf{\small Accuracy}\\
\hline
\small ynie/roberta-large-snli-mnli-fever-anli-R1-R2-R3-nli & \textbf{\small (0.57$\pm$0.04)} & \textbf{\small (0.56$\pm$0.02)} & \textbf{\small (0.58$\pm$0.02)} & \textbf{\small (0.72$\pm$0.03)} \\ 
\small nli-deberta-v3-base & \small (0.32$\pm$0.01) & \small (0.36$\pm$0.01) & \small (0.36$\pm$0.02) & \small (0.42$\pm$0.01) \\ 
\hline
\end{tabular}
\caption{{\bf Modified subject}: \small Fine-tuned NLI models' evaluation scores based on 4-entailment relations. \\
We repeated these experiments 5 times with different random seeds; we report averages and standard deviation, for  \\
8 epochs and a batch size of 32 data points.}
\label{tab:finetuned-subj}
\end{table*}
\begin{table*}
\begin{tabular}{lllll}
\hline
\textbf{\small NLI system} & \textbf{\small F1} & \textbf{\small Precision} & \textbf{\small Recall} & \textbf{\small Accuracy}\\
\hline
\small ynie/roberta-large-snli-mnli-fever-anli-R1-R2-R3-nli & \textbf{\small (0.55$\pm$0.08)} & \textbf{\small (0.54$\pm$0.02)} & \textbf{\small (0.58$\pm$0.02)} & \textbf{ \small (0.71$\pm$0.03)} \\ 
\small nli-deberta-v3-base & \small (0.31$\pm$0.01) & \small (0.36$\pm$0.03) & \small (0.39$\pm$0.02) & \small (0.38$\pm$0.02) \\ 
\hline
\end{tabular}
\caption{{\bf Modified verb}: \small Fine-tuned NLI models' evaluation scores based on 4-entailment relations. \\
We repeated these experiments 5 times with different random seeds; we report averages and standard deviation, for  \\
8 epochs and a batch size of 32 data points.}
\label{tab:finetuned-verb}
\end{table*}
\begin{table*}
\begin{tabular}{lllll}
\hline
\textbf{\small NLI system} & \textbf{\small F1} & \textbf{\small Precision} & \textbf{\small Recall} & \textbf{\small Accuracy}\\
\hline

\small ynie/roberta-large-snli-mnli-fever-anli-R1-R2-R3-nli & \textbf{ \small (0.57$\pm$0.04)} & \textbf{\small (0.57$\pm$0.01)} & \textbf{\small (0.58 $\pm$ 0.02 )} & \textbf{\small (0.72$\pm$0.02)} \\ 
\small nli-deberta-v3-base & \small (0.38$\pm$0.02) & \small (0.41$\pm$0.02) & \small (0.41$\pm$0.02) & \small (0.49$\pm$0.02) \\ 
\hline
\end{tabular}
\caption{{\bf Modified object}: \small Fine-tuned NLI models' evaluation scores based on 4-entailment relations. \\
We repeated these experiments 5 times with different random seeds; we report averages and standard deviation, for  \\
8 epochs and a batch size of 32 data points.}
\label{tab:finetuned-obj}
\end{table*}

\end{document}